\documentclass[conference]{IEEEtran}
\IEEEoverridecommandlockouts
\usepackage{amsthm}
\usepackage{amsmath}
\usepackage{amssymb}
\usepackage{booktabs}
\usepackage{siunitx}
\usepackage{cite}
\usepackage{graphics, framed}
\usepackage{graphicx}
\usepackage{epsfig}
\usepackage{epstopdf}
\usepackage{subfigure}
\usepackage{url}
\usepackage{multimedia}
\usepackage{paralist}
\usepackage[printonlyused]{acronym}
\usepackage{units}
\usepackage{balance}
\usepackage{multirow}
\newcommand{\FGR}[1]{Fig.~\ref{#1}}

\newcommand{\SEC}[1]{Section~\ref{#1}}
\newcommand{\TAB}[1]{Table~\ref{#1}}

\renewcommand{\baselinestretch}{0.94}
\acrodef{CFAR}[CFAR]{constant false alarm rate}
\acrodef{AWGN}[AWGN]{additive white Gaussian noise}
\acrodef{CR}[CR]{cognitive radio}
\acrodef{SNR}[SNR]{signal--to--noise ratio}
\acrodef{ADAM}[ADAM]{adaptive moment estimation}
\acrodef{ISM}[ISM]{industrial, scientific and medical}
\acrodef{CPU}[CPU]{central processing unit}
\acrodef{GPU}[GPU]{graphics processing unit}
\acrodef{LB}[LB]{likelihood based}
\acrodef{FB}[FB]{feature based}
\acrodef{AM}[AM]{amplitude modulation}
\acrodef{PSK}[PSK]{phase shift keying}
\acrodef{PAM}[PAM]{pulse amplitude modulation}
\acrodef{FSK}[FSK]{frequency shift keying}
\acrodef{QAM}[QAM]{quadrature amplitude modulation}
\acrodef{I/Q}[I/Q]{in--phase/quadrature}
\acrodef{AI}[AI]{artificial intelligence}
\acrodef{CNN}[CNN]{convolutional neural network}
\acrodef{ResNet}[ResNet]{residual network}
\acrodef{LSTM}[LSTM]{long short term memory}
\acrodef{DL}[DL]{deep learning}
\acrodef{CLDNN}[CLDNN]{convolutional long short term memory fully connected deep neural network}
\acrodef{5G}[5G]{5\textsuperscript{th}--Generation}
\acrodef{BW}[BW]{bandwidth}
\acrodef{CW}[CW]{continuous wave}
\acrodef{D2D}[D2D]{device--to--device}
\acrodef{M2M}[M2M]{machine--to--machine}
\acrodef{dB}[dB]{decibel}
\acrodef{dBi}[dBi]{decibel isotropic}
\acrodef{dBm}[dBm]{decibel over a milliwatt}
\acrodef{Gbps}[Gbps]{Gigabit per second}
\acrodef{GHz}[GHz]{gigahertz}
\acrodef{Hz}[Hz]{hertz}
\acrodef{IF}[IF]{intermediate frequency}
\acrodef{IFFT}[IFFT]{inverse fast Fourier Transform}
\acrodef{LO}[LO]{local oscillator}
\acrodef{LOS}[LOS]{line--of--sight}
\acrodef{MHz}[MHz]{megahertz}
\acrodef{MIMO}[MIMO]{multiple--input multiple--output}
\acrodef{mmWave}[mmWave]{millimeter wave}
\acrodef{NGWN}[NGWN]{next generation wireless network}
\acrodef{NLOS}[NLOS]{non line--of--sight}
\acrodef{QoS}[QoS]{quality of service}
\acrodef{RF}[RF]{radio frequency}
\acrodef{MLE}[MLE]{maximum likelihood estimation}
\acrodef{UHF}[UHF]{ultra high frequency}
\acrodef{AMC}[AMC]{Automatic modulation classification}
\acrodef{SCF}[SCF]{spectral correlation function}
\acrodef{SVM}[SVM]{support vector machine}
\acrodef{FFT}[FFT]{fast Fourier Transform}
\acrodef{RNN}[RNN]{recurrent neural network} 
\acrodef{ReLU}[ReLU]{rectified linear unit}
\acrodef{PCA}[PCA]{principal component analysis}
\ifCLASSINFOpdf
\else
\fi

\begin{document}
\title{Robust and Fast Automatic Modulation Classification with CNN under Multipath Fading Channels\thanks{This paper has been accepted for the presentation in the 2020 IEEE 91st Vehicular Technology Conference (VTC2020-Spring).}}
\IEEEoverridecommandlockouts 

\author{\IEEEauthorblockN{K{\"{u}}r{\c{s}}at Tekb{\i}y{\i}k\IEEEauthorrefmark{1}\IEEEauthorrefmark{2}, Ali R{\i}za Ekti\IEEEauthorrefmark{1}\IEEEauthorrefmark{4}, Ali G\"{o}r\c{c}in\IEEEauthorrefmark{1}\IEEEauthorrefmark{5}, G{\"{u}}ne{\c{s}} Karabulut Kurt\IEEEauthorrefmark{2}, Cihat Ke\c{c}eci\IEEEauthorrefmark{1}\IEEEauthorrefmark{3}}
\IEEEauthorblockA{\IEEEauthorrefmark{1}Informatics and Information Security Research Center (B{\.{I}}LGEM), T{\"{U}}B{\.{I}}TAK, Kocaeli, Turkey}

\IEEEauthorblockA{\IEEEauthorrefmark{2}Department of Electronics and Communication Engineering, Istanbul Technical University, {\.{I}}stanbul, Turkey}

\IEEEauthorblockA{\IEEEauthorrefmark{4}Department of Electrical--Electronics Engineering, Bal{{\i}}kesir University, Bal{{\i}}kesir, Turkey}

\IEEEauthorblockA{\IEEEauthorrefmark{5}Faculty of Electronics and Communications Engineering, Y{{\i}}ld{{\i}}z Technical University, {\.{I}}stanbul, Turkey}

\IEEEauthorblockA{\IEEEauthorrefmark{3}Department of Electrical--Electronics Engineering, Bo\u{g}azi\c{c}i University, {\.{I}}stanbul, Turkey\\ Emails: \texttt{\{kursat.tekbiyik, cihat.kececi\}@tubitak.gov.tr,} \texttt{arekti@balikesir.edu.tr,}\\ \texttt{agorcin@yildiz.edu.tr,} \texttt{gkurt@itu.edu.tr}}}

\maketitle

\begin{abstract}
\ac{AMC} has been studied for more than a quarter of a century; however, it has been difficult to design a classifier that operates successfully under changing multipath fading conditions and other impairments. Recently, \ac{DL}--based methods are adopted by \ac{AMC} systems and major improvements are reported. In this paper, a novel \ac{CNN} classifier model is proposed to classify modulation classes in terms of their families, \textit{i.e.}, types. The proposed classifier is robust against realistic wireless channel impairments and in relation to that, when the data sets that are utilized for testing and evaluating the proposed methods are considered, it is seen that RadioML2016.10a is the main dataset utilized for testing and evaluation of the proposed methods. However, the channel effects incorporated in this dataset and some others may lack the appropriate modeling of the real--world conditions since it only considers two distributions for channel models for a single tap configuration. Therefore, in this paper, a more comprehensive dataset, named as HisarMod2019.1, is also introduced, considering real-life applicability. HisarMod2019.1 includes $26$ modulation classes passing through the channels with $5$ different fading types and several number of taps for classification.  It is shown that the proposed model performs better than the existing models in terms of both accuracy and training time under more realistic conditions. Even more, surpassed their performance when the RadioML2016.10a dataset is utilized.
\end{abstract}
\IEEEpeerreviewmaketitle
\acresetall

\section{Introduction}
\ac{AMC} has been considered as an important part of various military and civilian communication systems, such as electronic warfare, radio surveillance and spectrum awareness. As known, classical signal identification methods used in the past are based on complex collections of feature extraction methods, such as cyclostationarity, high--order cumulants and complex hierarchical decision trees. Furthermore, it should be noted that classical methods cannot be generalized over all signal types and they suffer from dynamic nature of the propagation channel and cannot be adopted easily if a new wireless communication technology emerges. On the other hand, \ac{DL} has been proposed as a useful method for such classification problems and recently have been applied to this domain intensively. However, these methods should also provide strong performance against the wireless impairments in that particular domain thus, robust \ac{AMC} methods based on \ac{DL} techniques should be investigated to achieve dependable, efficient and resilient classification performance under realistic wireless communication channel conditions. 
\subsection{Related Work}

Signal identification systems often use \ac{LB} and \ac{FB} techniques. Although, \ac{LB} methods make the probability of correct classification maximum, they suffer from high computational complexity. Also, they are not robust to model mismatches, such as channel coefficient estimates and timing offsets \cite{panagiotou2000likelihood, hameed2009likelihood, xu2010software}.

 On the other hand, in \ac{FB} approaches, it is required to find a feature which can distinguish the signal from others. However, single feature mostly is not sufficient to classify signals in a large set. In literature, the higher order statistics, wavelet transform, and cyclic characteristics are mainly proposed features for signal identification. For instance, the wavelet transform is utilized in the identification of \ac{FSK} and \ac{PSK} signals \cite{hong1999identification}. The higher order statistics such as higher order cumulants and moments which are another feature used in \ac{AMC} \cite{liu2006novel, swami2000hierarchical}. In addition to these features, \cite{nandi1995automatic} utilizes instantaneous amplitude, phase and frequency statistics in order to make modulation classification. Howbeit, it is explicitly known that these features hamper to perform well in real--world conditions such as multipath channel fading, frequency, and timing offsets. Although the most powerful \ac{FB} approach, cyclostationarity--based features are resistant to mismatches compared to other features \cite{dobre2015signal}, it suffers from high computational complexity.
 
Machine learning--based approaches have been recently adopted to \ac{AMC}. For example, \ac{CNN}, \ac{CLDNN} and \ac{LSTM} can be said as the most popular deep neural network architectures for \ac{AMC}. \cite{kulin2018end} proposes using \ac{CNN} with \ac{I/Q} data and \ac{FFT} for \ac{AMC} and interference identification in \ac{ISM} band. It is shown that \acp{RNN} can be utilized for \ac{AMC} under Rayleigh channel with uncertain noise condition \cite{hu2018robust}. In addition to proposing \ac{CLDNN} for \ac{AMC}, \cite{ramjee2019fast} compares it to other existing models under different subsampling rates and different number of samples. Furthermore, it aims to reduce training time for online learning by utilizing subsampling and \ac{PCA}. \ac{LSTM} is proposed in  \cite{rajendran2018deep}, but it does not allow online learning and has long enough training time to require very high computing capacity. The RadioML2016.10a dataset\footnote{\scriptsize It is available on http://opendata.deepsig.io/datasets/2016.10/RML2016.10a.tar.bz2} \cite{o2016radio} is widely used in the literature. However, a system that works under real--conditions should be designed to operate under different channel conditions. Due to the dynamic nature of propagation channel and severe multipath effects, the existing available datasets cannot fulfill to provide the desired real--world conditions. RadioML2018.01a introduced in \cite{o2018over} includes over--the--air recordings of $24$ digital and analog modulation types. However, it cannot provide information about the channel parameters since this data set is based on measurement. Therefore, this dataset cannot allow generating information about how the channel conditions affect the performance of the model trained on the dataset. Furthermore, it has not serious diversity because it is created in the laboratory environment where there is no significant change in the channel parameters such as fading and number of taps. In this case, there is a need for a data set that includes both actual channel conditions and controlled channel parameters. It is also necessary to design a \ac{DL} model that can work under real channel conditions.
\subsection{Contributions}

The main contributions of this study are two fold and can be summarized as follows:
\begin{itemize}
    \item  First, aforementioned discussions show that currently, there is no comprehensive, inclusive, and controlled dataset that integrates the severe multipath effects for the real--world channel conditions. Therefore, we first introduce a new and more challenging modulation dataset, HisarMod2019.1 \cite{hisarmod}. This new public dataset provides wireless signals under ideal, static, Rayleigh, Rician ($k = 3$), and Nakagami--m ($m = 2$) channel conditions with various numbers of channel taps. Thus, it becomes possible to observe more realistic channel conditions for the proposed \ac{DL}--based \ac{AMC} methods.
    
    \item  More importantly, a new \ac{CNN} model with optimal performance in terms of accuracy and training time under more realistic conditions is proposed. The proposed method exhibits higher performance under both in HisarMod2019.1 dataset and existing RadioML2016.10a dataset when compared to the available classifiers. The new CNN consists of four convolution and two dense layers. In addition to its high performance, the model has lower training complexity when compared to the available techniques, thus, the training process is relatively short.

\end{itemize}

\section{HisarMod2019.1: A New Dataset}\label{sec:dataset}

In order to increase the diversity in signal datasets, we create a new dataset called as HisarMod2019.1, which includes $26$ classes and $5$ different modulation families passing through $5$ different wireless communication channel. During the generation of the dataset, MATLAB 2017a is employed for creating random bit sequences, symbols, and wireless fading channels. 

The dataset includes $26$ modulation types from $5$ different modulation families which are analog, \ac{FSK}, \ac{PAM}, \ac{PSK}, and \ac{QAM}. All modulation types are listed in \TAB{tab:hisar_mod_types}. In the dataset, there are $1500$ signals, which have the length of $1024$ \ac{I/Q} samples, for each modulation type. To make HisarMod2019.1 similar to RadioML2016.10a for fair comparison, there are $20$ different \ac{SNR} levels in between \unit{-20}{dB} and \unit{18}{dB}. As a result, the dataset covers totally $780000$ signals. When generating signals, oversampling rate is chosen as $2$ and raised cosine pulse shaping filter is employed with roll--off factor of $0.35$.

Furthermore, the dataset consists of signals passing through $5$ different wireless communication channels which are ideal, static, Rayleigh, Rician ($k=3$), and Nakagami--m ($m=2$). These channels are equally likely distributed over the dataset; therefore, there are 300 signals for each modulation type and each \ac{SNR} level. Ideal channel refers that there is no fading, but \ac{AWGN}. In the static channel, the channel coefficients are randomly determined at the beginning and they remain constant over the propagation time. The signals passing through Rayleigh channel are employed to make the system resistant against \ac{NLOS} conditions. On the other hand, Rician fading with shape parameter, $k$, of $3$ is utilized owing to the fact that the dataset covers a mild fading. In addition to these channel models, the distribution of received power is selected as Nakagami--m with shape parameter, $m$, of $2$ for the rest of the signals in the dataset. As a result, the dataset includes signals with different fading models. Noting that the number of multipath channel taps are equally likely selected as $4$ and $6$ which are adopted from ITU--R M1225 \cite{ITURM1225}.

\begin{table}[!t]
\centering
\caption{HisarMod2019.1 includes 26 different modulation types from 5 different modulation families.}
\begin{tabular}{c l}
\toprule
Modulation Family       & Modulation Types \\ \midrule
\multirow{6}{*}{Analog} & AM--DSB           \\
                        & AM--SC            \\
                        & AM--USB           \\
                        & AM--LSB           \\
                        & FM               \\
                        & PM               \\ \midrule
\multirow{4}{*}{FSK}    & 2--FSK            \\
                        & 4--FSK            \\
                        & 8--FSK            \\
                        & 16--FSK           \\ \midrule
\multirow{3}{*}{PAM}    & 4--PAM            \\
                        & 8--PAM            \\
                        & 16--PAM           \\ \midrule
\multirow{6}{*}{PSK}    & BPSK             \\
                        & QPSK             \\
                        & 8--PSK            \\
                        & 16--PSK           \\
                        & 32--PSK           \\
                        & 64--PSK           \\ \midrule
\multirow{7}{*}{QAM}    & 4--QAM            \\
                        & 8--QAM            \\
                        & 16--QAM           \\
                        & 32--QAM           \\
                        & 64--QAM           \\
                        & 128--QAM          \\
                        & 256--QAM          \\ \bottomrule
\end{tabular}
\label{tab:hisar_mod_types}
\end{table}

\section{The Proposed \ac{CNN} Model}\label{sec:cnn_model}

\begin{figure*}[!t]
    \centering
    \includegraphics[width=\textwidth]{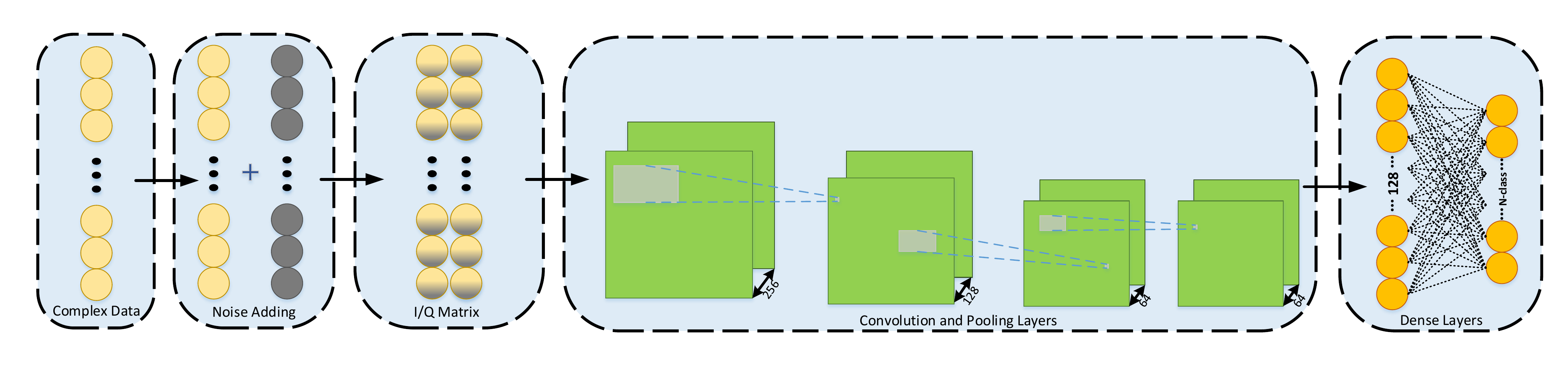}
    \caption{The proposed \ac{CNN} model consists of four convolution and pooling layers and two dense layers.}
    \label{fig:cnn_model}
\end{figure*}

In this paper, a \ac{CNN} model is built by using Keras which is an open source machine learning library \cite{chollet2015keras}. The proposed \ac{CNN} model involves four convolution and pooling layers terminated by two dense layers. The \ac{ReLU} activation function, which is defined as
\begin{align}
    x_{out} = \max(0,\,\omega x_{in} + b),
    \label{eq:relu}
\end{align}
is employed in each convolution layer. In \eqref{eq:relu}, $x_{in}$, $x_{out}$, $\omega$, and $b$ are the input and output of the function, weight, and bias, respectively.  In this model, it is chosen that the model gets narrower in terms of the number of filters in each convolution layer through the end of the feature extraction part of the model. Our experience with many different configurations indicated that the models that get narrower in each following convolutional layer provides better results in terms of classification and reduce training time. Indeed, for the optimal performance, we employed $256$ filters in the first layer while the last layer had $64$ filters.  The first dense layer is formed by $128$ neurons and \ac{ReLU} activation function. The dense layer is followed by a softmax activation function which computes the probabilities for each class as
\begin{align}
    S\left(y_{i}\right)=\frac{e^{y_{i}}}{\sum_{j} e^{y_{j}}}, \; i,\,j = 1, 2,\cdots, N,
\end{align}
where $y_{i}$ and $N$ are any element of classes and the number of classes, respectively. Moreover, the \ac{ADAM} optimizer is used to estimate the model parameters with the learning rate of $10^{-4}$. The \ac{CNN} model architecture is depicted in \FGR{fig:cnn_model}. Furthermore, the layout for the proposed \ac{CNN} model is given in \TAB{tab:cnn_layout}. During the training process, we use early stopping to terminate the process if the validation loss converges to a level enough. As a result, the model is preserved to be overfitted. As seen in \FGR{fig:cnn_model}, there is a layer, which adds noise at each epoch; thus, it also prevents the model to overfit. The power of noise is determined according to the desired \ac{SNR} level.

In the training and test stages, we employ four NVIDIA Tesla V100 \acp{GPU} by operating them in parallel. It is seen that the proposed \ac{CNN} model is too light compared to \ac{CLDNN} \cite{ramjee2019fast} and \ac{LSTM} \cite{rajendran2018deep}. For example, the proposed \ac{CNN} model has $15$ million trainable parameters, whereas \ac{CLDNN} has $27$ million trainable parameters for HisarMod2019.1 dataset. Furthermore, \ac{CNN} model takes one--quarter time of \ac{LSTM} per epoch.

\begin{table}[!t]
\centering
\caption{The proposed CNN layout for the proposed dataset HisarMod2019.1 and RadioML2016.10a.}
\begin{tabular}{c c c}
\toprule
\multirow{2}{*}{Layer} & \multicolumn{2}{c}{Output Dimensions} \\
                       & HisarMod2019.1                     & RadioML2016.10a                 \\ \midrule
Input                  & $2\times1024$                & $2\times128$               \\
Noise Layer            & $2\times1024$                & --                          \\
Conv1                  & $2\times1024\times256$       & $2\times128\times256$                           \\
Max\_Pool1                  & $2\times512\times256$   & $2\times64\times256$                           \\
Dropout1                    & $2\times512\times256$   & $2\times64\times256$                           \\
Conv2                  & $2\times512\times128$        & $2\times64\times128$                           \\
Max\_Pool2                & $2\times256\times128$     & $2\times32\times128$                          \\
Dropout2                  & $2\times256\times128$     & $2\times32\times128$                         \\
Conv3                  & $2\times256\times64$         & $2\times32\times64$                          \\
Max\_Pool3            & $2\times128\times64$          & $2\times16\times64$                          \\
Dropout3                  & $2\times128\times64$      & $2\times16\times64$                           \\
Conv4                  & $2\times128\times64$         & $2\times16\times64$                           \\
Max\_Pool4                  & $2\times64\times64$     & $2\times8\times64$                           \\
Dropout4                  & $2\times64\times64$       & $2\times8\times64$                           \\
Flatten                 & $8192$                      & $1024$ \\
Dense1                & $128$                            & $128$ \\
Dense2                 & $5$                        & $10$  \\
Trainable Par.         & $15,764,53$                   & $6,595,94$  \\                    \bottomrule
\end{tabular}
\label{tab:cnn_layout}
\end{table}

\section{Classification Results}\label{sec:results}

The proposed model is tested in both the HisarMod2019.1 and the RadioML2016.10a datasets. The test results are provided below.

\subsection{HisarMod2019.1 Dataset Classification Results}
\begin{figure}[!t]
    \centering
    \includegraphics[width=\linewidth]{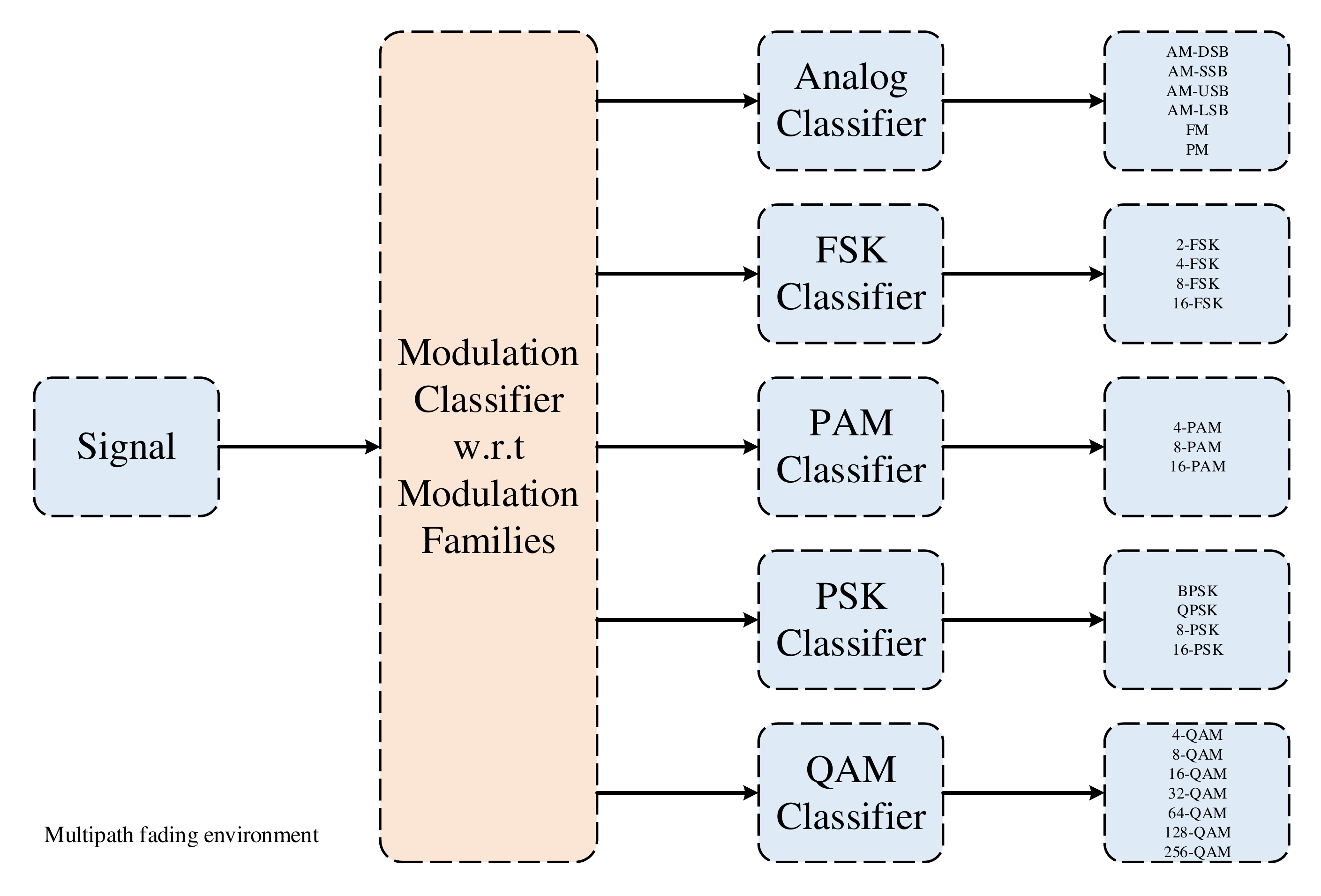}
    \caption{In the multipath fading environment, it is not easy to deal with a large dataset; hence, it can be handled in two steps: modulation family classification, and modulation type classification.}
    \label{fig:hierarchical_model}
\end{figure}

\begin{figure*}[!t]
    \centering
    \subfigure[]{
    \label{fig:hisarmod_cnn_cldnn}
    \includegraphics[width=0.45\linewidth]{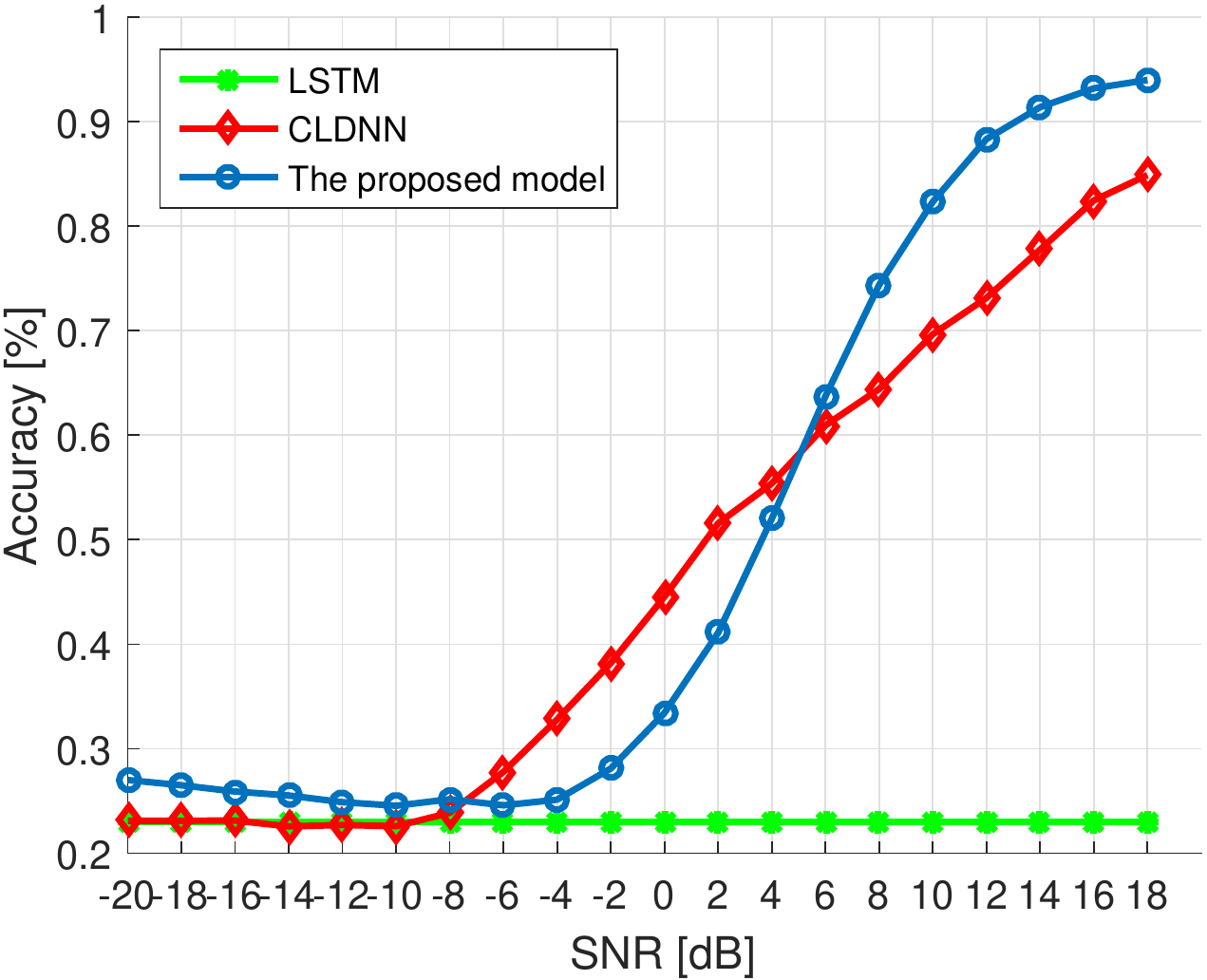}
    }
    \qquad
    \subfigure[]{
    \label{fig:radioml_cnn_cldnn}
    \includegraphics[width=0.45\linewidth]{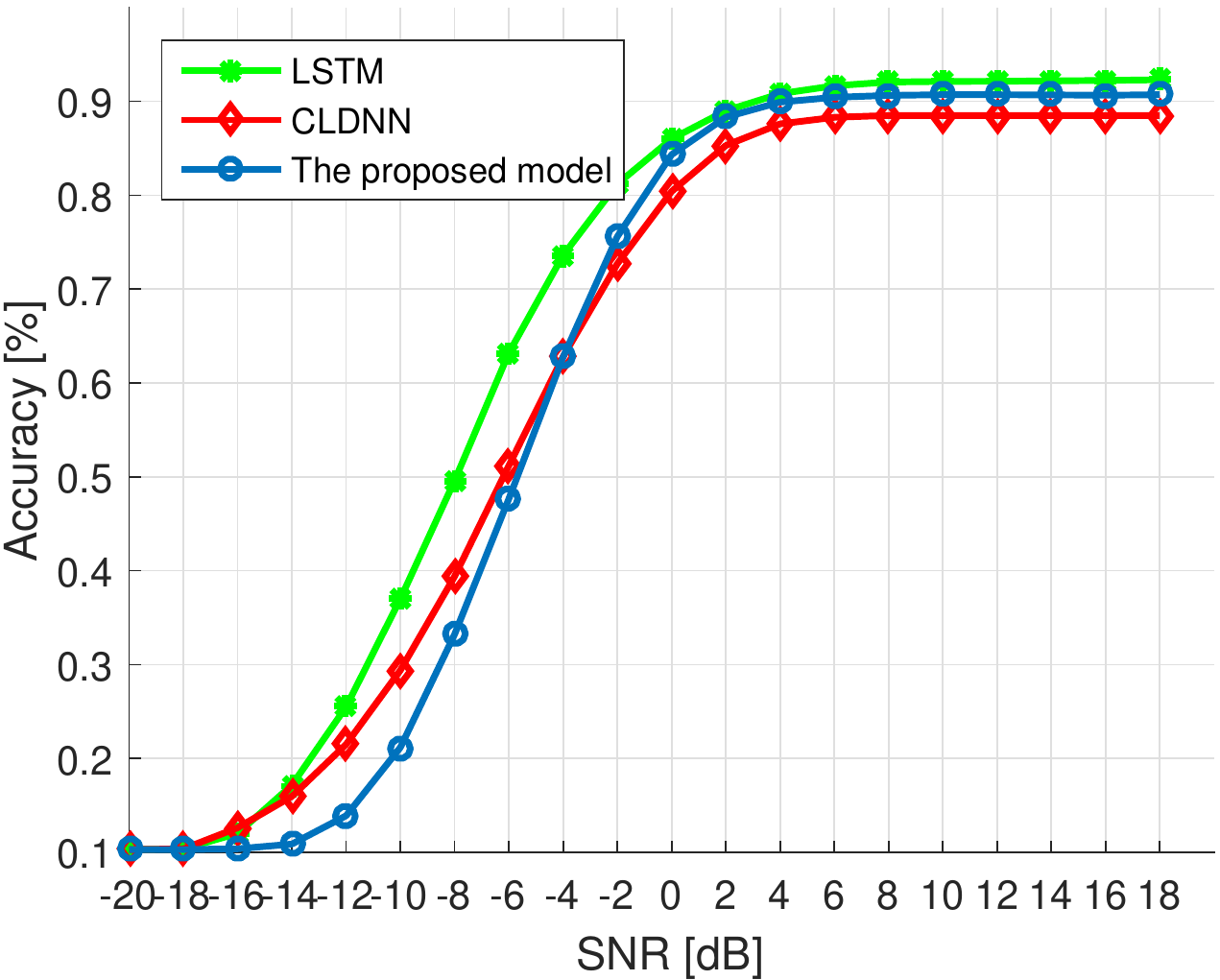}
    }
    \caption{The accuracy values for \ac{LSTM}, \ac{CLDNN} and the proposed \ac{CNN} models in (a) the HisarMod2019.1, (b) RadioML2016.10a datasets.}
    \label{fig:acc_snr}
\end{figure*}

As detailed in \SEC{sec:dataset}, the HisarMod2019.1 covers $26$ different modulation types. It is not that easy to handle so many signal types in the fading environment. It is expected that they are confused each other due to the deterioration in their amplitude and phase. Thus, in this study, we use an approach like the data binning method by labeling signals with respect to their modulation families such as analog, \ac{FSK}, \ac{PAM}, \ac{PSK}, and \ac{QAM}. The hierarchical approach is depicted in \FGR{fig:hierarchical_model}. Firstly, we aim to classify signals in terms of modulation families. Then, each modulation type can be identified in the family subset. One should keep in mind that this study focuses on the classification of the modulation families not the order of each modulation type for the HisarMod2019.1 dataset. The dataset is split as $8/15$, $2/15$, and $5/15$ for training, validation, and test sets, respectively.

As stated before, the early stopping is employed in the training stage. The first layer of the \ac{CNN} adds noise to data according to the \ac{SNR} level. As a result, the model becomes more robust to overfitting.

The model gives meaningful results at \ac{SNR} levels higher than $2$ dB. It might be said that the model makes a random choice between modulation families at low \ac{SNR} values. Considering the nature of wireless communications, the model performs well for the expected \ac{SNR} values. The dataset is also employed with the \ac{CLDNN} model. It is noted that we employ the \ac{CLDNN} and \ac{LSTM} models as detailed in \cite{ramjee2019fast} and \cite{rajendran2018deep} without any adjustment. Also, the proposed \ac{CNN} model shows better performance than the existing \ac{CLDNN} and \ac{LSTM} models in HisarMod2019.1 dataset. For example, it exceeds $80\%$ accuracy at \unit{8}{dB} \ac{SNR}; however, \ac{CLDNN} performs with the same accuracy at \unit{16}{dB} \ac{SNR}. While \ac{CLDNN} does not achieve $90\%$ accuracy, our model exceeds this level at \unit{14}{dB} and higher. The maximum accuracy values for the proposed \ac{CNN} model and state of the art \ac{CLDNN} model are $94\%$ and $85\%$, respectively. Surprisingly, \ac{LSTM} cannot show acceptable classification results; however, it performs well in RadioML2016.10a. At \ac{SNR} values, the results are not meaningful in terms the classification accuracy since the false alarm rate gets higher. \FGR{fig:hisarmod_cnn_cldnn} denotes the accuracy values for \ac{CNN}, \ac{CLDNN}, and \ac{LSTM} models at the \ac{SNR} values in between [\unit{-20}{dB}, \unit{18}{dB}]. \FGR{fig:hisarmod_cnn_conf_matx} and \FGR{fig:hisarmod_cldnn_conf_matx} show the confusion matrices for the proposed \ac{CNN} model and \ac{CLDNN} model, respectively. Both of them have difficulty in the identification of \ac{QAM} signals. On the other hand, \ac{LSTM} recognizes signals as analog modulated signals regardless of the received signal type. Hence, the confusion matrices are not provided for the \ac{LSTM} model.

\begin{figure*}[!t]
    \centering
    \subfigure[]{%
        \label{fig:hisarmod_cnn_0dB}%
        \includegraphics[width=0.21\textwidth]{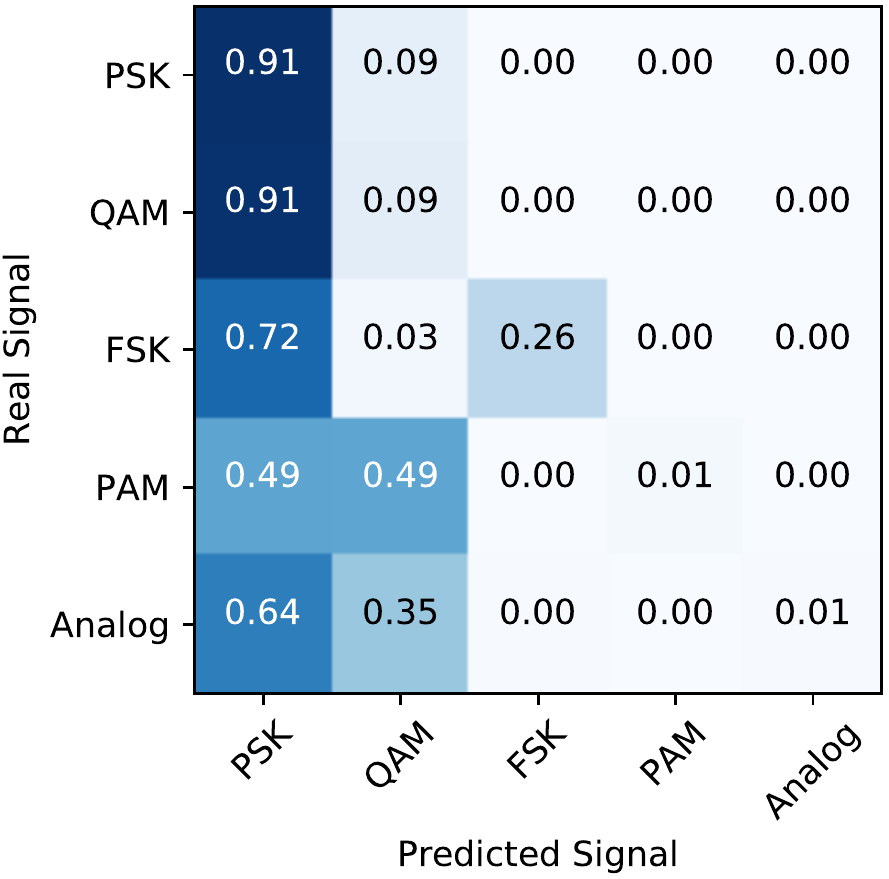}}%
    \qquad
    \subfigure[]{%
        \label{fig:hisarmod_cnn_6dB}%
        \includegraphics[width=0.21\textwidth]{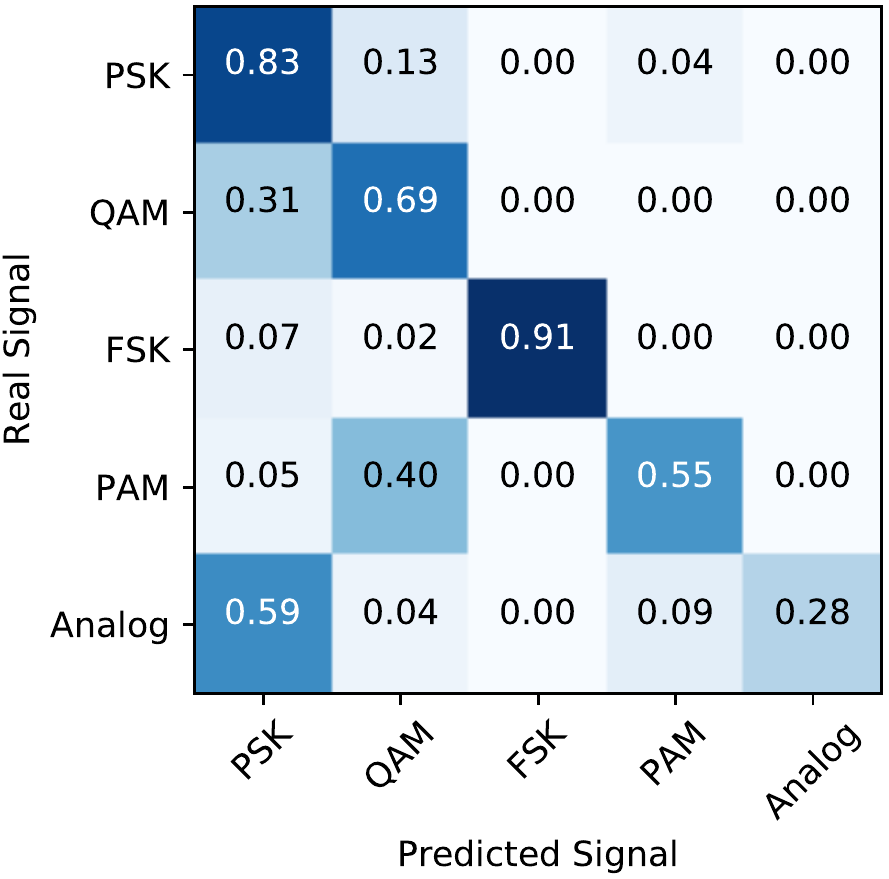}}
    \qquad
    \subfigure[]{%
        \label{fig:hisarmod_cnn_12dB}%
        \includegraphics[width=0.21\textwidth]{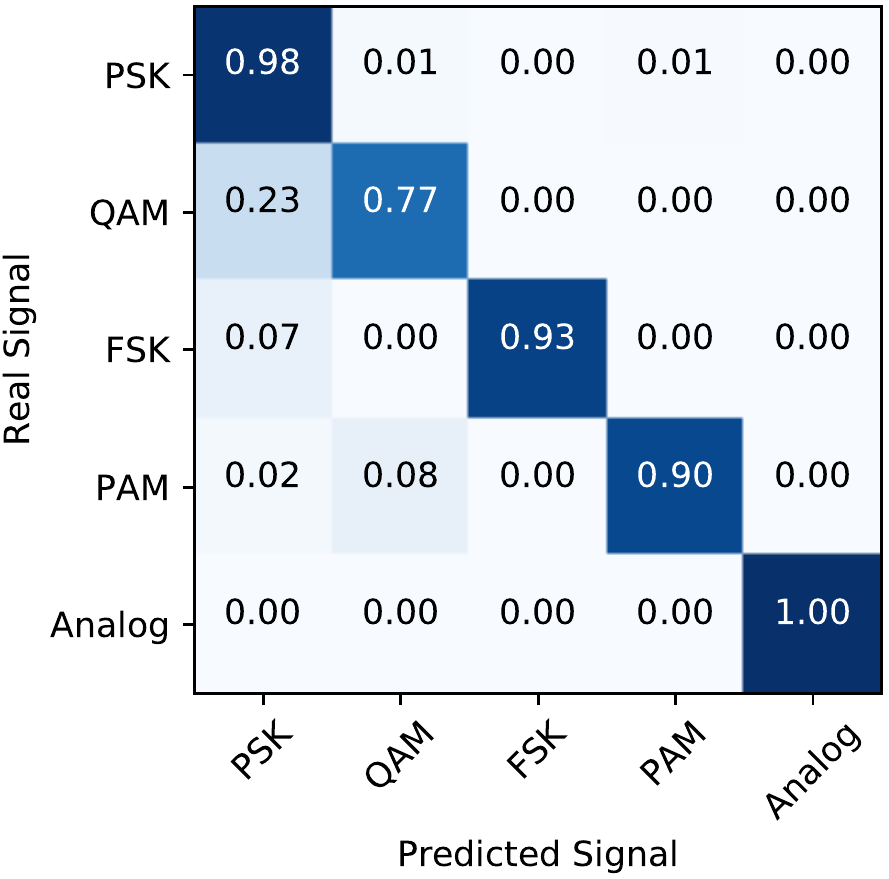}}%
     \qquad
    \subfigure[]{%
        \label{fig:hisarmod_cnn_18dB}%
        \includegraphics[width=0.21\textwidth]{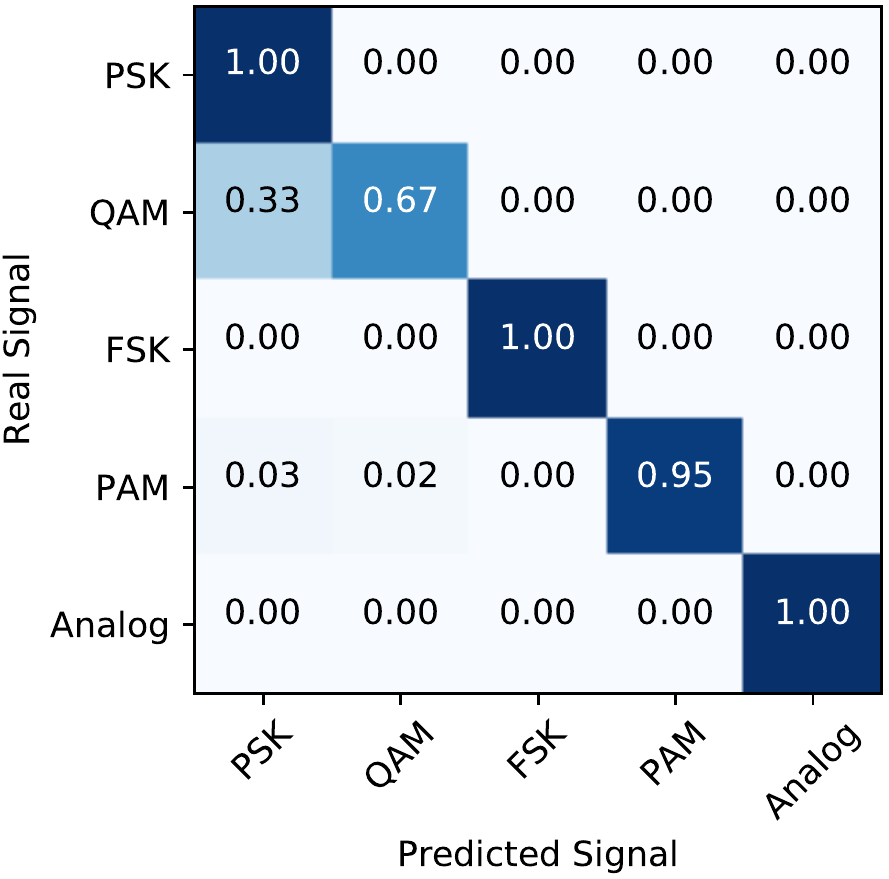}}%
    \caption{The confusion matrices of the proposed \ac{CNN} model test results at (a) \unit{0}{dB}, (b) \unit{6}{dB}, (c) \unit{12}{dB}, (d) \unit{18}{dB}, when the HisarMod2019.1 dataset is used.}
    \label{fig:hisarmod_cnn_conf_matx}
\end{figure*}

\begin{figure*}[!t]
    \centering
    \subfigure[]{%
        \label{fig:hisarmod_cldnn_0dB}%
        \includegraphics[width=0.21\textwidth]{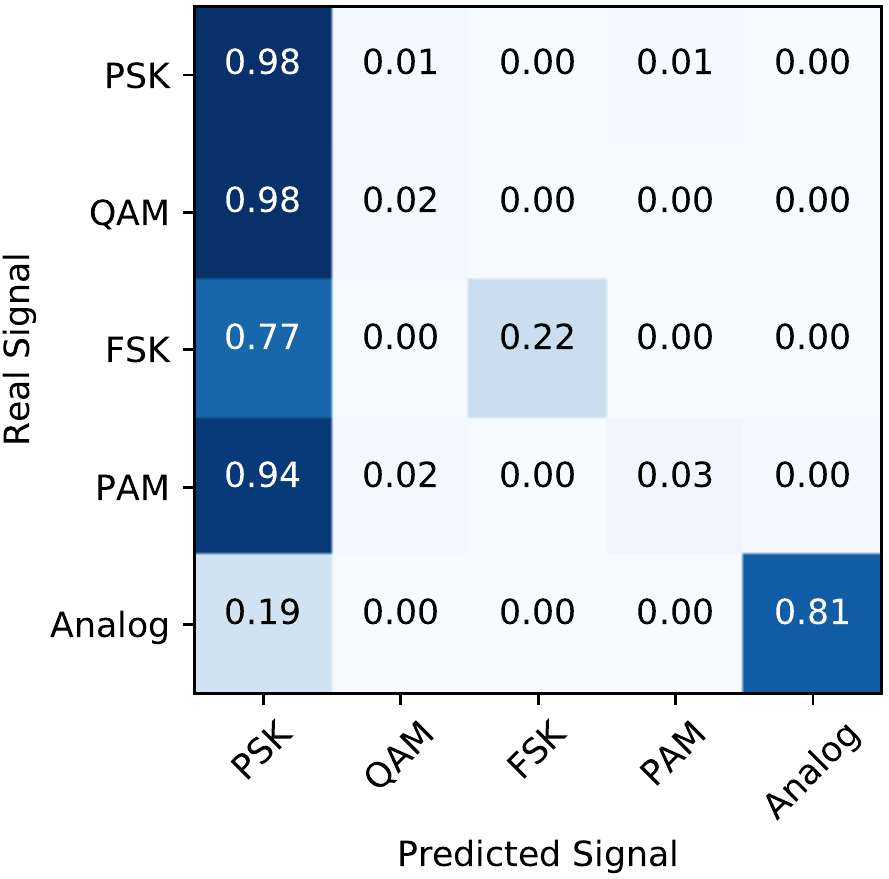}}%
    \qquad
    \subfigure[]{%
        \label{fig:hisarmod_cldnn_6dB}%
        \includegraphics[width=0.21\textwidth]{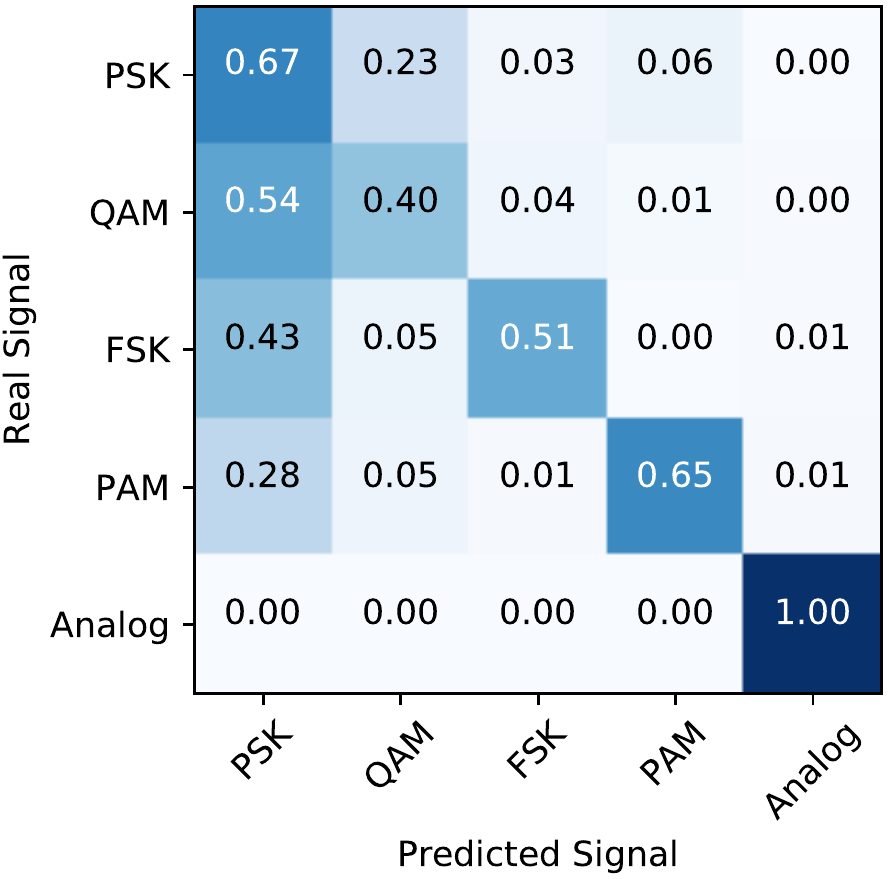}}
    \qquad
    \subfigure[]{%
        \label{fig:hisarmod_cldnn_12dB}%
        \includegraphics[width=0.21\textwidth]{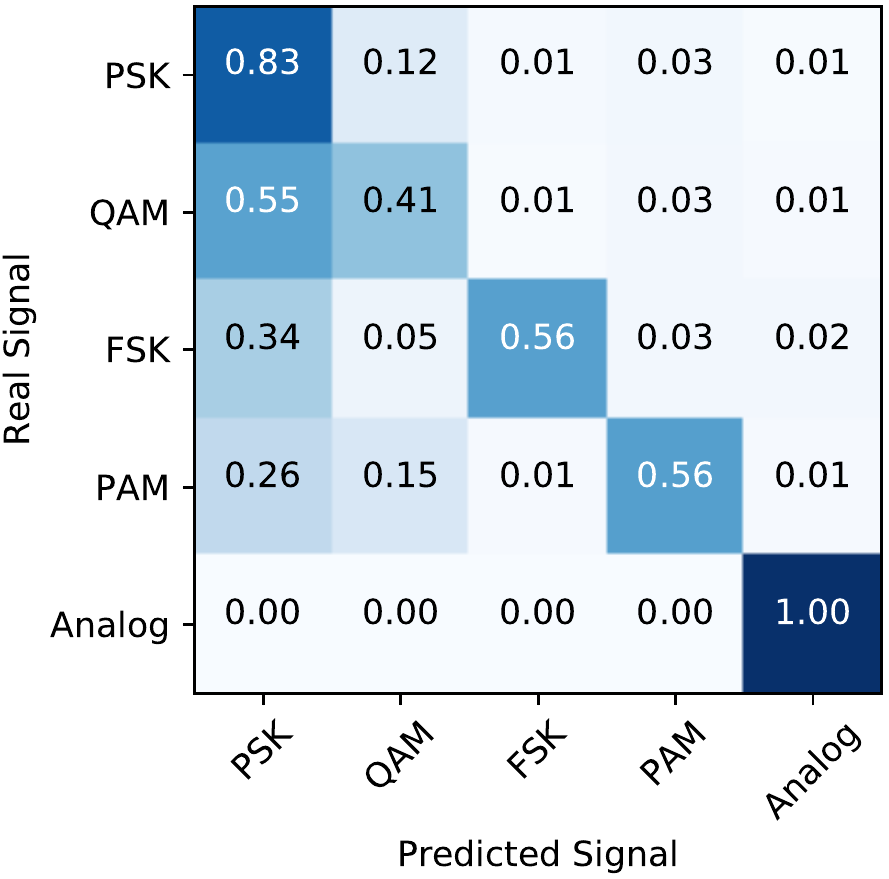}}%
     \qquad
    \subfigure[]{%
        \label{fig:hisarmod_cldnn_18dB}%
        \includegraphics[width=0.21\textwidth]{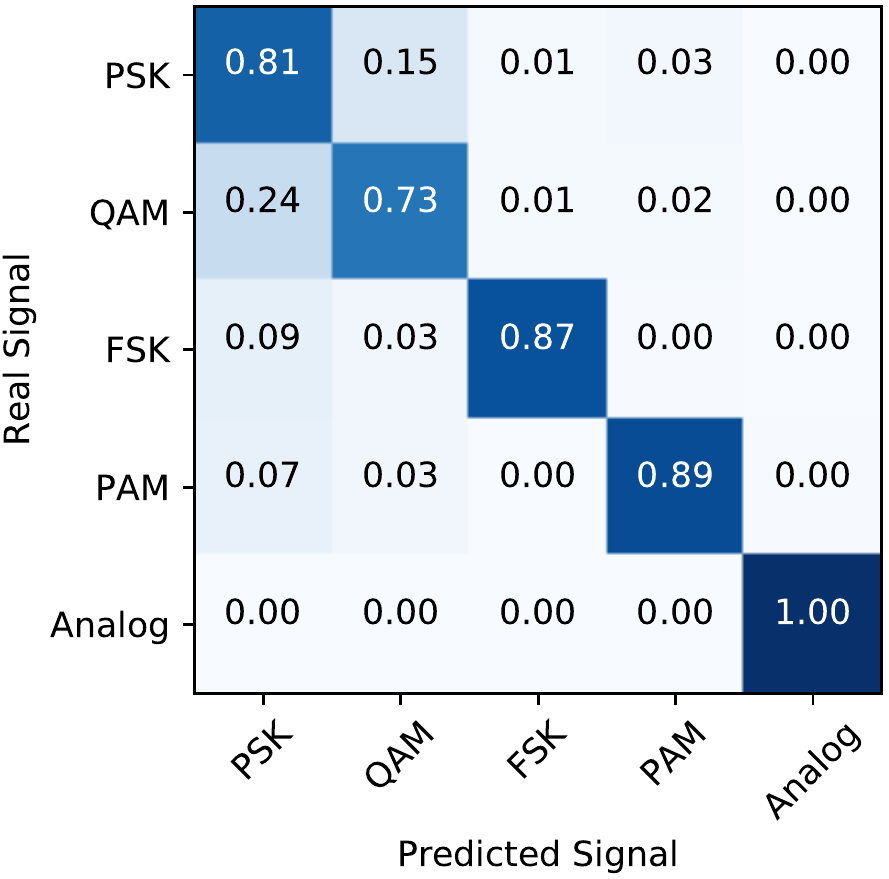}}%
    \caption{The confusion matrices of the \ac{CLDNN} model test results at (a) \unit{0}{dB}, (b) \unit{6}{dB}, (c) \unit{12}{dB}, (d) \unit{18}{dB}, when the HisarMod2019.1 dataset is used.}
    \label{fig:hisarmod_cldnn_conf_matx}
\end{figure*}

\subsection{RadioML2016.10a Dataset Classification Results}
RadioML dataset is heavily used in modulation classification studies and it is a well accepted dataset by the literature. Therefore, in order to show the robustness of our proposed \ac{CNN} model, we also test our model in RadioML dataset to observe its performance. In this section, RadioML2016.10a dataset is employed. It consists of synthetic signals with $10$ modulation types. The modulation types covered by the dataset are listed as: AM--DSB, WBFM, GFSK, CPFSK, 4--PAM, BPSK, QPSK, 8--PSK, 16--QAM, and 64--QAM. Details for the generation and packaging of the dataset can be found in \cite{o2016radio}.

Here, the dataset is split into two parts (i.e. training and test) with equal number of signals. After training procedure, the models are tested with the rest of the signals. According to test results, the proposed \ac{CNN} model shows higher performance than the \ac{CLDNN} model at the \ac{SNR} levels higher than \unit{-2}{dB}. \ac{LSTM} performs slightly better than \ac{CNN}. The \ac{CLDNN} is able to reach the maximum accuracy of $88.5\%$. On the other hand, the proposed \ac{CNN} model performs with the maximum accuracy of $90.7\%$ even though it is not originally designed for the RadioML2016.10a dataset. Although \ac{LSTM} reaches up to $92.3\%$ accuracy, its computational complexity is extremely high. \FGR{fig:radioml_cnn_cldnn} denotes the accuracy values with respect to \ac{SNR} levels. The confusion matrices for the classification results of the proposed \ac{CNN} model are depicted in \FGR{fig:radioml_cnn_conf_matx}. It is observed that the model recognizes almost all signals as 8--PSK at low \ac{SNR} levels. \FGR{fig:radioml_cldnn_-6dB} shows the confusion matrix of the minimum \ac{SNR} value of which the model performs over $50\%$ accuracy. As can be seen from \FGR{fig:radioml_cldnn_-6dB}, the model gives poor results in modulation types other than 4--PAM. The proposed model achieves very high performance in all modulation types, except WBFM at \unit{6}{dB} and above.

Initial observations suggest that the proposed model can work with high performance both in a diverse dataset, HisarMod2019.1, and RadioML2016.10a which is a frequently used dataset.

\begin{figure*}[!t]
    \centering
    \subfigure[]{%
        \label{fig:radioml_cldnn_-12dB}%
        \includegraphics[width=0.41\textwidth]{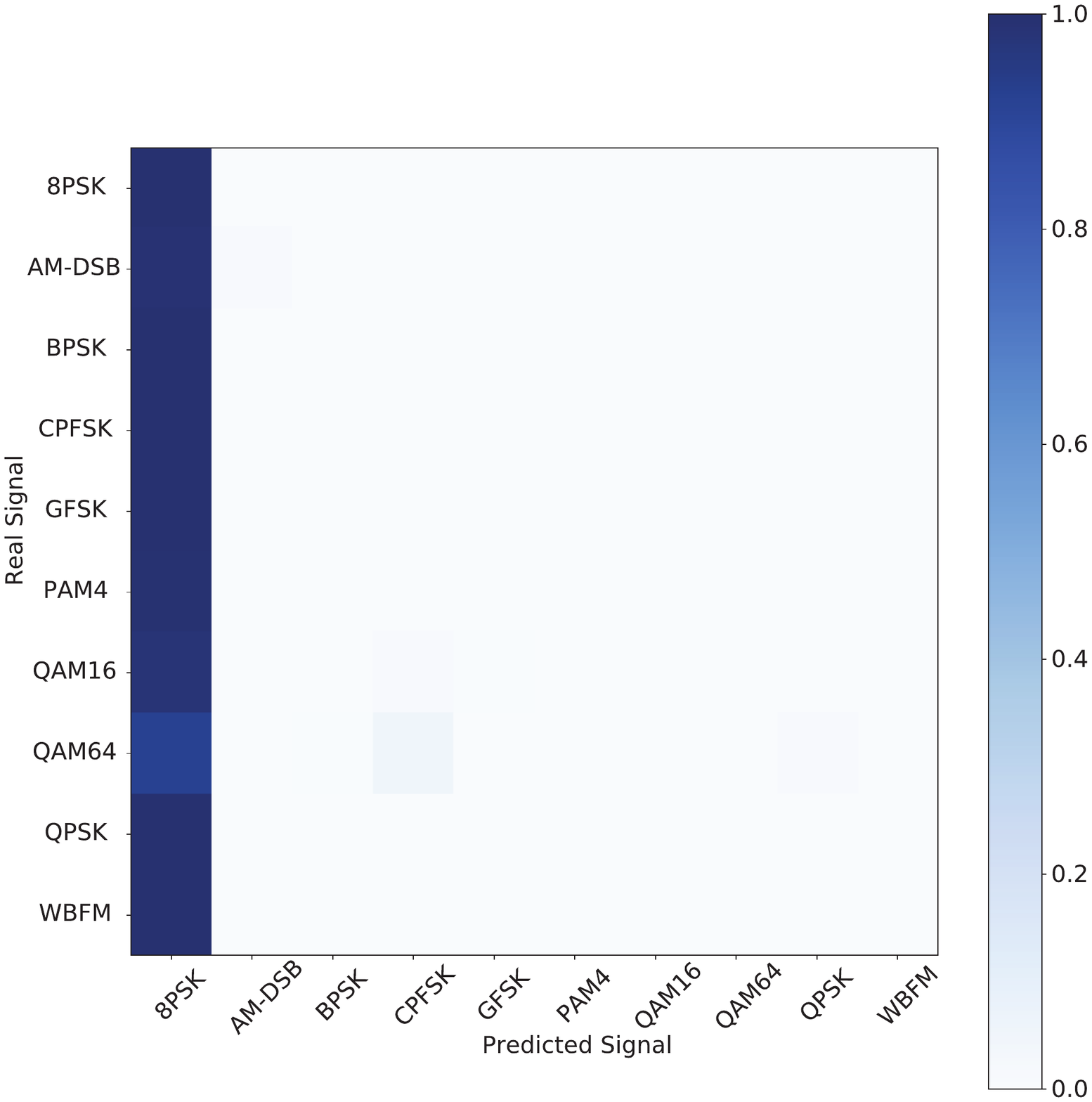}}%
    \qquad
    \subfigure[]{%
        \label{fig:radioml_cldnn_-6dB}%
        \includegraphics[width=0.41\textwidth]{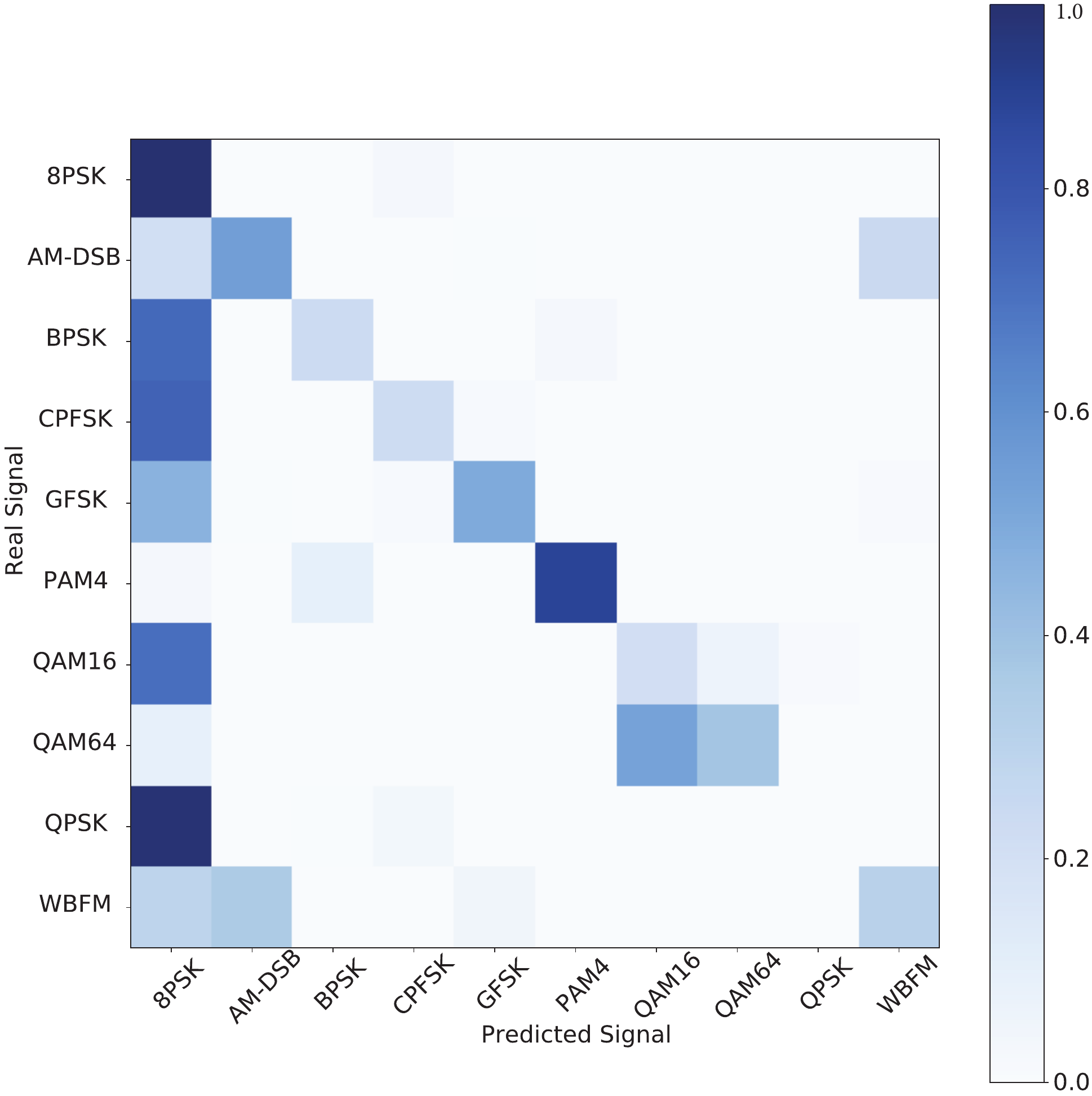}}
    \qquad
    \subfigure[]{%
        \label{fig:radioml_cldnn_0dB}%
        \includegraphics[width=0.41\textwidth]{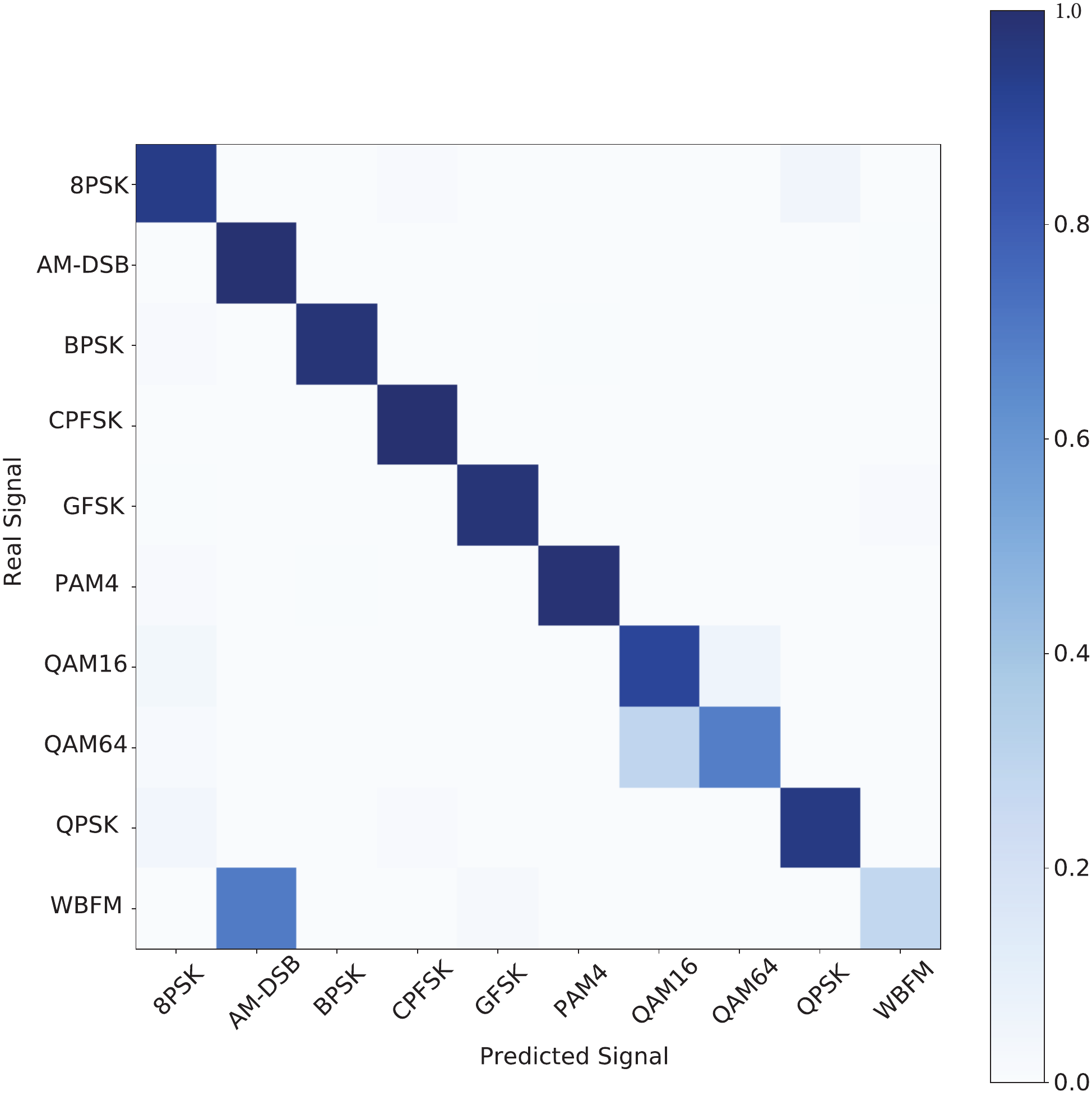}}%
     \qquad
    \subfigure[]{%
        \label{fig:radioml_cldnn_6dB}%
        \includegraphics[width=0.41\textwidth]{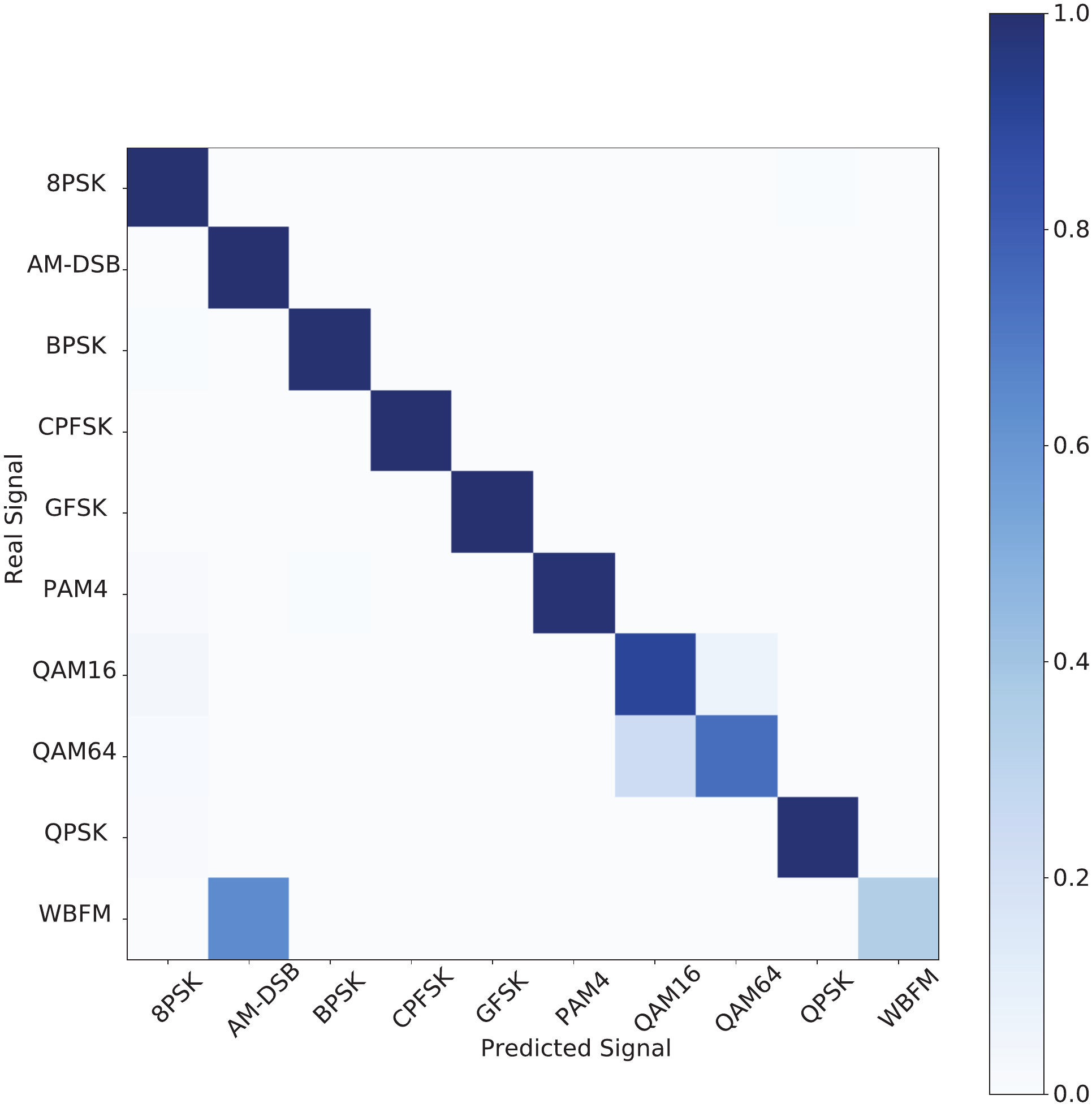}}%
    \caption{The confusion matrices of the proposed \ac{CNN} model test results at (a) \unit{-12}{dB}, (b) \unit{-6}{dB}, (c) \unit{0}{dB}, (d) \unit{6}{dB}, when the RadioML2016.10a dataset is used.}
    \label{fig:radioml_cnn_conf_matx}
\end{figure*}

\section{Concluding Remarks}\label{sec:conclusion}
In this study, we present a diverse new dataset, which consists of multipath fading signals with different number of channel taps, and a \ac{CNN} model for \ac{AMC}. The first stage of hierarchical classification architecture, which is the classification of modulation families, is realized with the proposed \ac{CNN} model on this dataset and the results compared with the \ac{CLDNN} model proposed in the literature. The results show that the proposed \ac{CNN} model performs significantly better than \ac{CLDNN}. Furthermore, the performance of the proposed \ac{CNN} model on the RadioML2016.10a dataset is examined. It is demonstrated that the proposed \ac{CNN} model is both faster and more accurate than the \ac{CLDNN} model. As a future work, we will investigate the classification of modulation orders assuming that the modulation family is identified. Finally, extensive search conducted for optimal model in this study shows that starting with an extensive set of filters, and then reducing their numbers  down step by step provides better results in terms of accuracy. This phenomenon will be investigated thoroughly and technical discussions will be provided in terms of explainable AI terminology.    

\section*{Acknowledgement}
This publication was made possible by NPRP12S-0225-190152 from the Qatar National Research Fund (a member of The Qatar Foundation). The statements made herein are solely the responsibility of the author[s].
\balance
\bibliographystyle{IEEEtran}
\bibliography{VTC2020_Modulation_Identification}
\end{document}